\documentclass[]{youtu} 

\usepackage{diagbox} 
\usepackage{multirow} 
\usepackage{booktabs} 

\usepackage{mathpazo}
\usepackage{graphicx}
\usepackage{natbib}
\usepackage{multirow}
\usepackage{subcaption}
\usepackage{lipsum} 
\usepackage{threeparttable} 
\newcommand\blfootnote[1]{%
  \begingroup
  \renewcommand\thefootnote{}\footnote{#1}%
  \addtocounter{footnote}{-1}%
  \endgroup
}
\title{ASPD: Unlocking Adaptive Serial-Parallel Decoding by Exploring Intrinsic Parallelism in LLMs}
\author{Youtu Team}
\affiliation{Full author list in contributions}

\date{August 5, 2025}
\correspondence{Ruizhi.Qiao@tencent.com}

\begin{document}

\abstract{

The increasing scale and complexity of large language models (LLMs) pose significant inference
latency challenges, primarily due to their autoregressive decoding paradigm characterized by the
sequential nature of next-token prediction. By re-examining the outputs of autoregressive models, we
observed that some segments exhibit parallelizable structures, which we term intrinsic parallelism.
Decoding each parallelizable branch simultaneously (\textit{i.e}. parallel decoding) can significantly improve
the overall inference speed of LLMs. In this paper, we propose an \textbf{A}daptive \textbf{S}erial-\textbf{P}arallel \textbf{D}ecoding (\textbf{ASPD}), which addresses two core challenges: automated construction of parallelizable
data and efficient parallel decoding mechanism. More specifically, we introduce a non-invasive
pipeline that automatically extracts and validates parallelizable structures from the responses of autoregressive
models. To empower efficient adaptive serial-parallel decoding, we implement a \textbf{Hybrid Decoding Engine} which enables seamless transitions between serial and parallel decoding modes while
maintaining a reusable KV cache, maximizing computational efficiency. Extensive evaluations across
General Tasks, Retrieval-Augmented Generation and Mathematical Reasoning demonstrate that \textbf{ASPD}
achieves unprecedented performance in both effectiveness and efficiency. Notably, on Vicuna Bench,
our method achieves up to 3.19x speedup (1.85x on average) while maintaining response quality
within 1\% difference compared to autoregressive models, realizing significant acceleration without
compromising generation quality. Our framework sets a groundbreaking benchmark for efficient
LLM parallel inference, paving the way for its deployment in latency-sensitive applications such as
AI-powered customer service bots and answer retrieval engines.
}
\maketitle

\vspace{-.1em}


\section{Introduction}

Recent advances in large language models (LLMs) have led to dramatic increases in both model size and context length. 
Although these enhancements improve the models' capabilities, they also introduce considerable inference latency, primarily due to the sequential nature of autoregressive decoding, presenting significant challenges for real-world applications.

\begin{figure}[htbp]
    \centering
    \includegraphics[width=\textwidth]{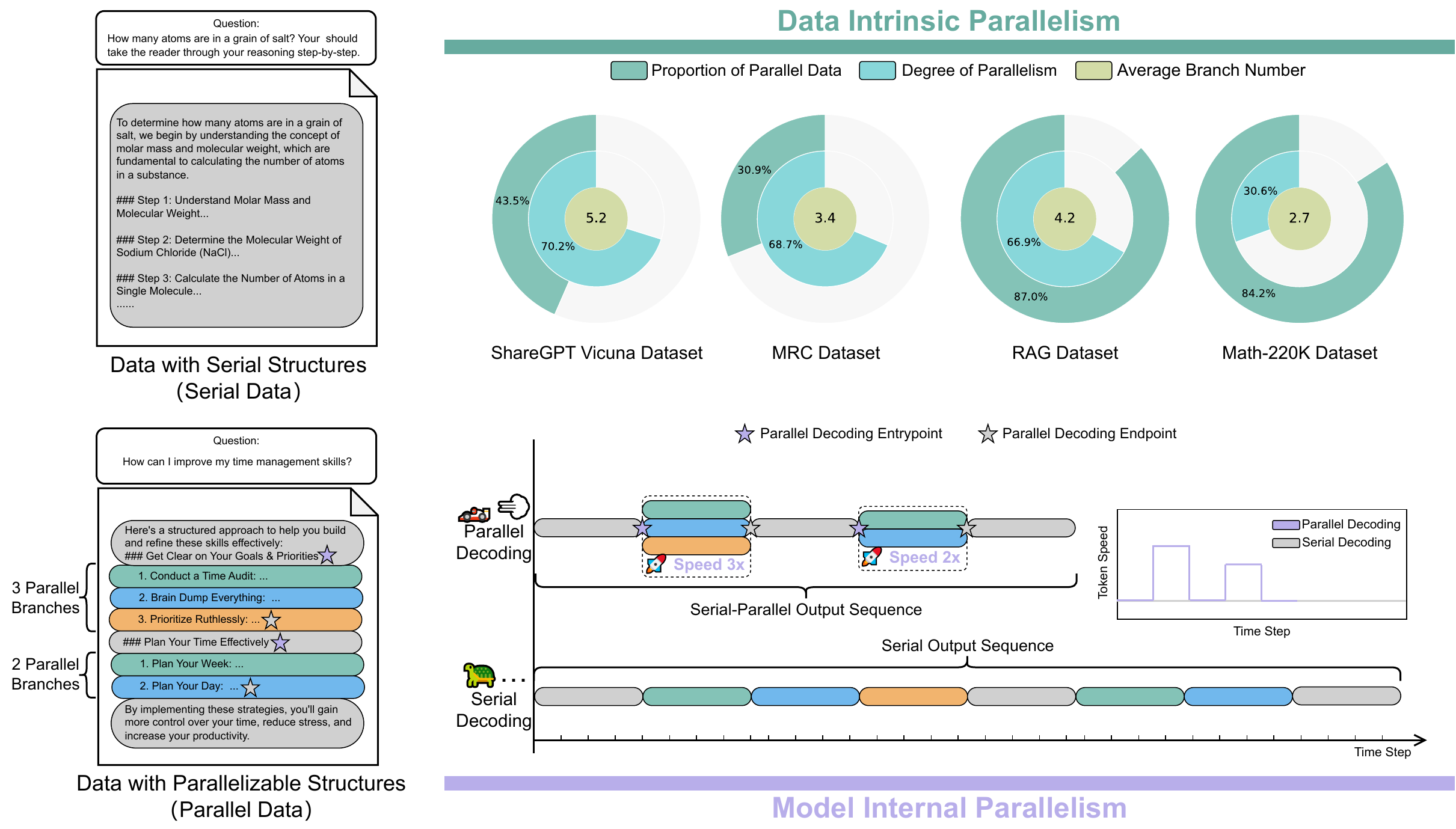}
    \caption{\textbf{Overview of Data Intrinsic and Model Internal Parallelism. }
    \textbf{\textit{Proportion of Parallel Data (PPD). }} The ratio of parallelizable samples in the benchmark;
    \textbf{\textit{Degree of Parallelism (DP). }} The proportion of parallel tokens among all tokens in the output;
    \textbf{\textit{Average Branch Number (ABN). }} The average number of parallel branches in the output.
    }
    \label{fig:ourmethod}
\end{figure}

Through extensive analysis of LLM outputs, we have made a crucial observation:
despite their sequential generation pattern, a substantial portion of the model's outputs exhibits inherent parallelism, which can be harnessed for concurrent processing.
As shown in Figure \ref{fig:ourmethod}, across various scenarios including general\protect\footnotemark[1], retrieval-augmented-generation\protect\footnotemark[2],
mathematical reasoning\protect\footnotemark[3], and our internal machine reading comprehension benchmarks, model responses consistently reveal substantial potential for parallelization.
Within these parallelizable components, different scenarios exhibit varying levels of parallelism, ranging from 30.6\% (Math) to 70.2\% (General).
By harnessing these inherently parallelizable responses for concurrent output, the model's response speed can be significantly enhanced.

\footnotetext[1]{\url{https://huggingface.co/datasets/anon8231489123/ShareGPT_Vicuna_unfiltered}}
\footnotetext[2]{\url{https://huggingface.co/datasets/neural-bridge/rag-dataset-12000}}
\footnotetext[3]{\url{https://huggingface.co/datasets/open-r1/OpenR1-Math-220k} }

However, this also raises several fundamental technical challenges:
First, identifying parallelizable segments while preserving semantic integrity is inherently difficult, due to the intricate dependencies present in natural language generation.
Second, it is critical to ensure strict independence across concurrent branches: Each must remain contextually isolated during decoding, yet collectively yield a coherent output when merged.
Third, the sophisticated coordination of positional information across concurrent branches presents significant architectural challenges,
particularly in preserving proper token relations and temporal coherence throughout concurrent generation.

Recent attempts to address these challenges have yielded promising but limited solutions. 
SoT \citep{ning2023skeleton} introduced a two-stage approach using response skeletons and parallel generation through external APIs, but this method incurs additional latency due to KV-cache limitations and increased memory overhead with batch processing.
APAR \citep{liu2024APAR} proposed discarding parallel branch's KV-caches during final generation, which, while addressing position encoding issues, compromises response quality.
PASTA \citep{jin2025learning} explored asynchronous parallel decoding with pre-allocated position ranges, but struggles with position encoding mismatches when actual generation lengths deviate from predictions.

Addressing these challenges is essential for unlocking the full potential of parallel decoding in practical LLM systems. To better exploit the intrinsic parallelism within autoregressive models, 
we present \textbf{A}daptive \textbf{S}erial-\textbf{P}arallel \textbf{D}ecoding, 
 a novel framework that more effectively harnesses the model's inherent parallel capabilities through a dual-perspective optimization of data utilization and architectural innovation. 
 Our approach first extracts inherent parallelism patterns from model responses to serve as training corpora for parallelization.
 We further propose an intra-response parallelization module that enables parallel processing in one go.
By implementing branch-specific attention masks and consistent positional indexing across parallel branches, 
our method ensures that: (1) parallel branch generation maintains behavioral consistency with native serial decoding from each branch's perspective; 
and (2) upon completion of all parallel branches, 
switching back to the primary branch incurs no information loss and recomputation overhead.
To achieve these objectives, we propose a \textbf{Hybrid Decoding Engine} that supports efficient parallel decoding and iterative serial-parallel decoding.

Our key contributions include:

1) \textbf{Non-invasive parallel data transformation pipeline. }
We develop an innovative pipeline that automatically discovers and extracts inherent parallelizable structures from autoregressive model responses,
which identifies semantically independent components that can be processed concurrently while preserving the response's original style.
This enables us to build high-quality parallel training corpora automatically without altering the response probability distribution.

2) \textbf{Internal parallelization module. }
Our framework introduces a custom branch-invisible mask that enables efficient parallel processing without the overhead of batching or threading, maintaining consistency with traditional autoregressive decoding patterns.

3) \textbf{Hybrid Decoding Engine supports efficient and cyclic serial-parallel decoding. }
We propose novel parallel positional encoding and training techniques that enable zero-overhead integration of parallel branches and support multiple rounds of parallel decoding.

4) \textbf{Comprehensive evaluation achieving an optimal balance between effectiveness and efficiency. }
Through extensive evaluation across diverse benchmarks - including general tasks (Vicuna Bench, MT Bench), retrieval-augmented generation (Neural-Bridge-RAG), 
and mathematical reasoning (MATH500, AMC23, GPQADiamond, AIME2024, AIME2025) 
- we demonstrate significant improvements in both effectiveness and efficiency compared to existing approaches.

\section{Related Work}
\label{sec:related}


\textbf{Parallelization Objectives and Paradigms. }
Existing research on parallelizing large language model inference broadly pursues two distinct objectives: 
improving response quality or accelerating generation speed. Quality-oriented parallelization techniques such as best-of-n sampling\citep{brown2024large}, beam search\citep{wiseman2016sequence},
self-consistency\citep{wang2022self}, majority voting\citep{chen2024more} , Monte Carlo tree search\citep{zhang2024accessing} and Tree of Thoughts \citep{yao2023tree} achieve better results through multiple parallel sampling iterations. These methods focus on test-time scaling without optimizing the sampling process efficiency.
Parallel scaling\citep{chen2025parallel} increases the model's parallel computation during both training and inference by applying multiple learnable transformations to the input.
In contrast to the above approaches, our work aligns with speed-oriented parallelization, 
targeting inherent parallelism within single responses by efficiently decomposing sequential generation into parallelizable units while maintaining coherence.
Our work focuses exclusively on the latter category, which we systematically classify below.

\begin{figure}[htb]
    \centering
    \begin{subfigure}[b]{0.94\textwidth}
        \centering
        \includegraphics[width=\textwidth]{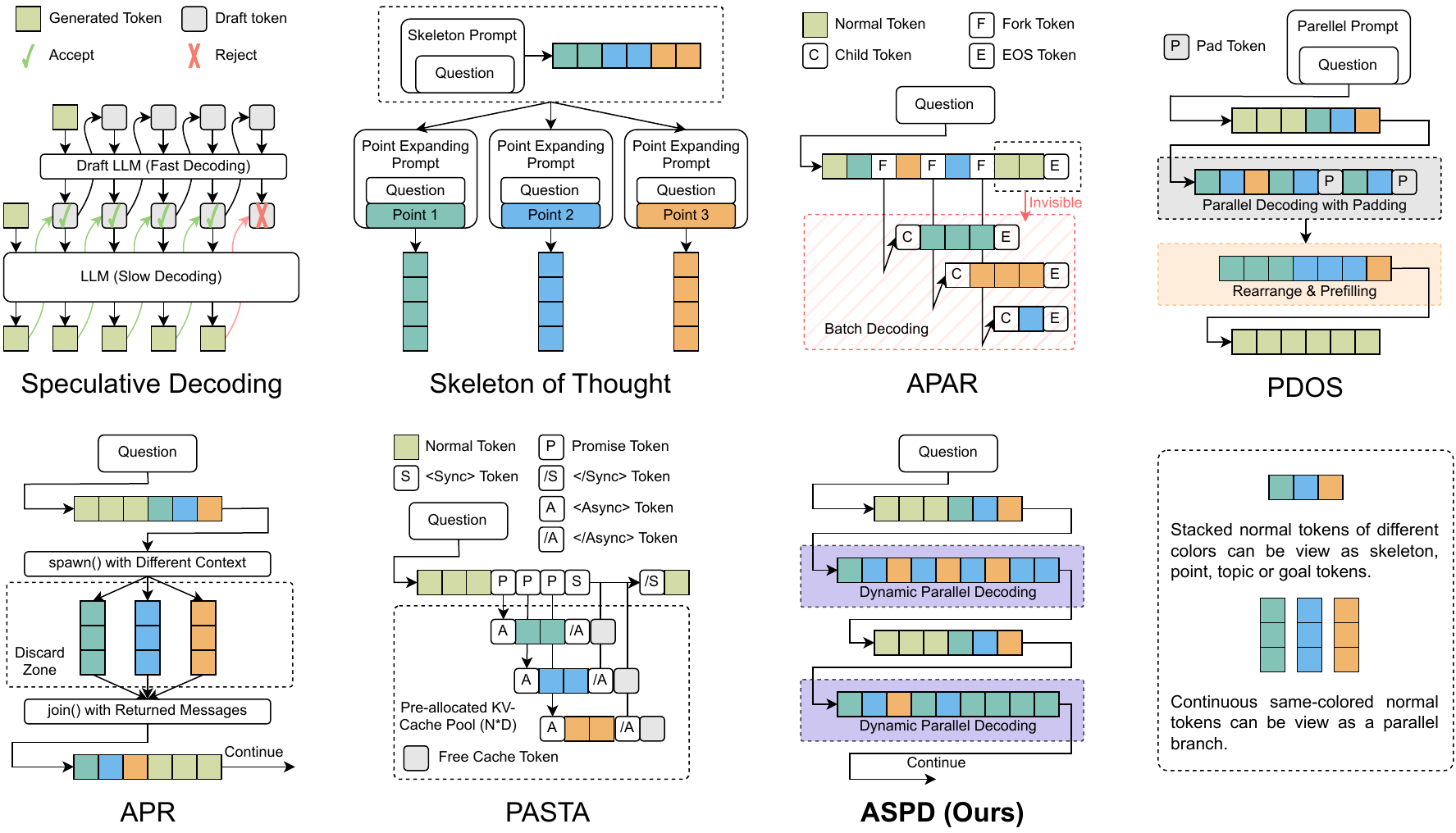}
        \caption{Overview of Parallel Decoding Methods}
        \label{fig:paralleldecoding_methods}
    \end{subfigure}
    \begin{subfigure}[b]{0.94\textwidth}
        \centering
        \includegraphics[width=\textwidth]{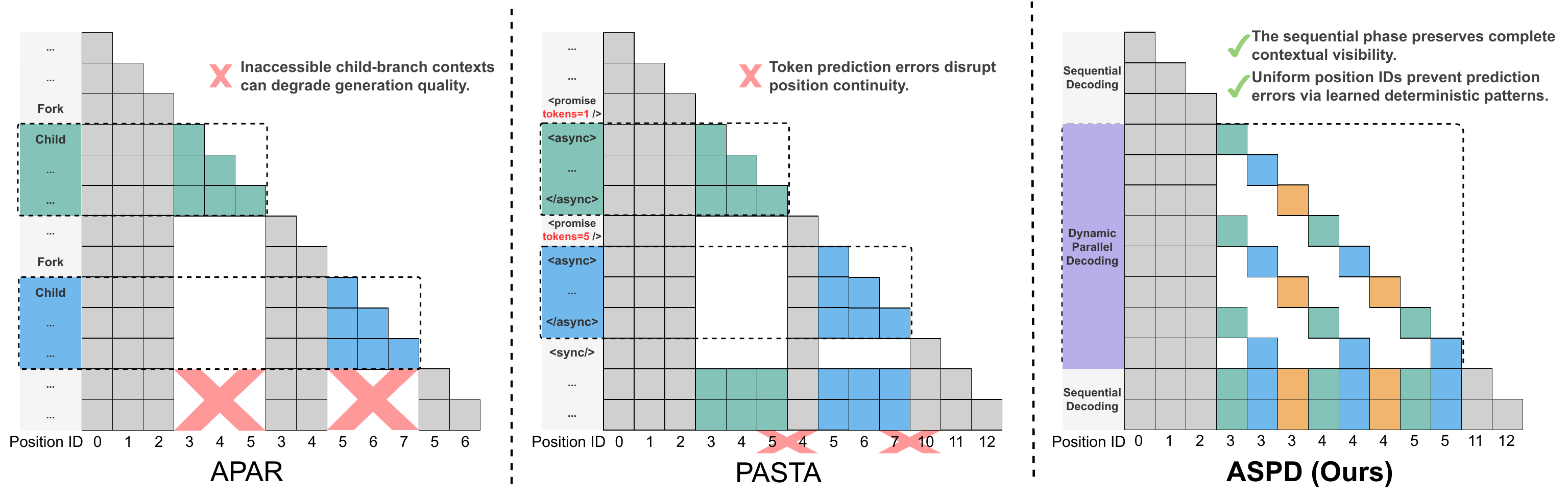}
        \caption{Internal Parallel Decoding Mechanisms. }
        \label{fig:position_mask}
    \end{subfigure}
    \caption{Related Parallel Decoding Methods.}
    \label{fig:all_methods}
\end{figure}

\subsection{Orthogonal Acceleration Techniques}
\textbf{Speculative Decoding. } Recent advances in speculative decoding \citep{leviathan2023fast} have emerged as a promising direction for accelerating LLM inference. 
For example, Medusa \citep{cai2023medusa} uses multiple prediction heads to generate several tokens at once. Eagle \citep{li2024eagle,li2024eagle2} and HASS \citep{zhang2024learning} employ a lightweight single-layer Transformer to predict the hidden state of the next token. SPACE \citep{yi2024generation} enables parallel token generation and validation through semi-autoregressive fine-tuning and automatic correction decoding algorithms. Self-speculative decoding \citep{zhang2023draft} skips intermediate layers selectively, using its own model to generate draft tokens. While effective, these approaches remain fundamentally \textit{sequential at the token level} as they must maintain the autoregressive property. 
Specifically, tokens can only be confirmed sequentially even if predicted in parallel, 
limiting their theoretical speedup. 
These methods complement rather than substitute intra-response parallelism, 
as they operate at different granularities - token-level versus segment-level parallelization.

\subsection{Parallelization via Prompt Engineering}
This category encompasses methods that leverage carefully crafted prompts to achieve parallel generation.
SoT \citep{ning2023skeleton} adopts a two-phase approach: first generating skeletal outlines (\textit{e.g.}, bullet points), 
followed by concurrent branch generation through batch decoding or API calls.
This introduces substantial overhead from KV-cache reinitialization during phase transitions.
Additionally, this approach suffers from increased memory overhead due to batch processing and compromised response quality from constrained prompt templates.
PDOS \citep{yu2025accelerate} advances this approach by implementing internal parallel decoding through specialized masks and logits processors,
eliminating the need for batching and multi-stage prompting.
Nevertheless, as a prompt-based solution, it encounters inherent limitations:
the model's parallel capabilities remain underutilized, and the necessity for content re-prefilling during mode transitions diminishes efficiency gains.

\subsection{Architecture-Modified Parallelization}
There have a few lines of work that modify attention mechanisms or training procedures to enable parallelism.

\textbf{Visible Branch Architectures. } Systems like \citep{hsu2025group} and \citep{rodionov2025hogwild} allow inter-branch communication but suffer from  backtracking costs (\textit{e.g.}, 
when branches generate overlapping content). 
These methods focus more on branch collaboration rather than accelerating through independent parallel branch decoding.\\

\textbf{Hidden Branch Architectures. } 
As shown in Figure \ref{fig:position_mask}, APAR \citep{liu2024APAR} , PASTA \citep{jin2025learning} and APR \citep{pan2025learning} enforce branch isolation but introduce critical trade-offs. 
\begin{itemize}
\item \textbf{\textit{Parallel Information Loss. } }
APAR discards parallel branch KV-caches during integration, degrading output coherence.
APR facilitates inter-model communication during transitions between serial and parallel decoding modes. 
However, it only shares abbreviated text summaries rather than complete KV-states, 
which  compromises contextual coherence and continuity.
\item \textbf{\textit{Positional Mismatch. } }
PASTA's pre-allocated position ranges cause encoding conflicts when actual parallel lengths deviate from predictions, which further impacts the response quality in subsequent serial phases.
\item \textbf{\textit{Insufficient Intrinsic Parallelism Exploitation. }}
APAR employs predefined rules and regular expressions to parallelize existing sequential responses, limiting its applicability to data with obvious formatting while struggling with mathematical and code content.
PASTA parallelizes sequential responses through model rewriting but lacks further validation of independence and completeness between parallel branches.
\end{itemize}
Our concurrent work Multiverse \citep{yang2025multiverse}, focusing on the Mathematical Reasoning task, also explores parallelism in large language model generation, 
implementing parallel branch generation through SGLang\citep{zheng2024sglang} and leveraging Radix Attention for KV cache reuse.
In contrast, our approach performs decoding within a single sequence, providing inherent continuity of the KV cache to directly reuse.

The proposed \textbf{ASPD} framework better harnesses intrinsic parallelism through a non-invasive data transformation pipeline.
By introducing branch-invisible masks and shared position encodings, we enable lossless and seamless transitions between serial and parallel decoding modes.
We systematically investigate diverse model architectures and inference paradigms 
, demonstrating significant acceleration effects without quality degradation across multiple 
domains, particularly in General Tasks and Retrieval-Augmented Generation scenarios.

\section{Methodology}
\begin{figure}[htbp]
    \centering
    \includegraphics[width=\textwidth]{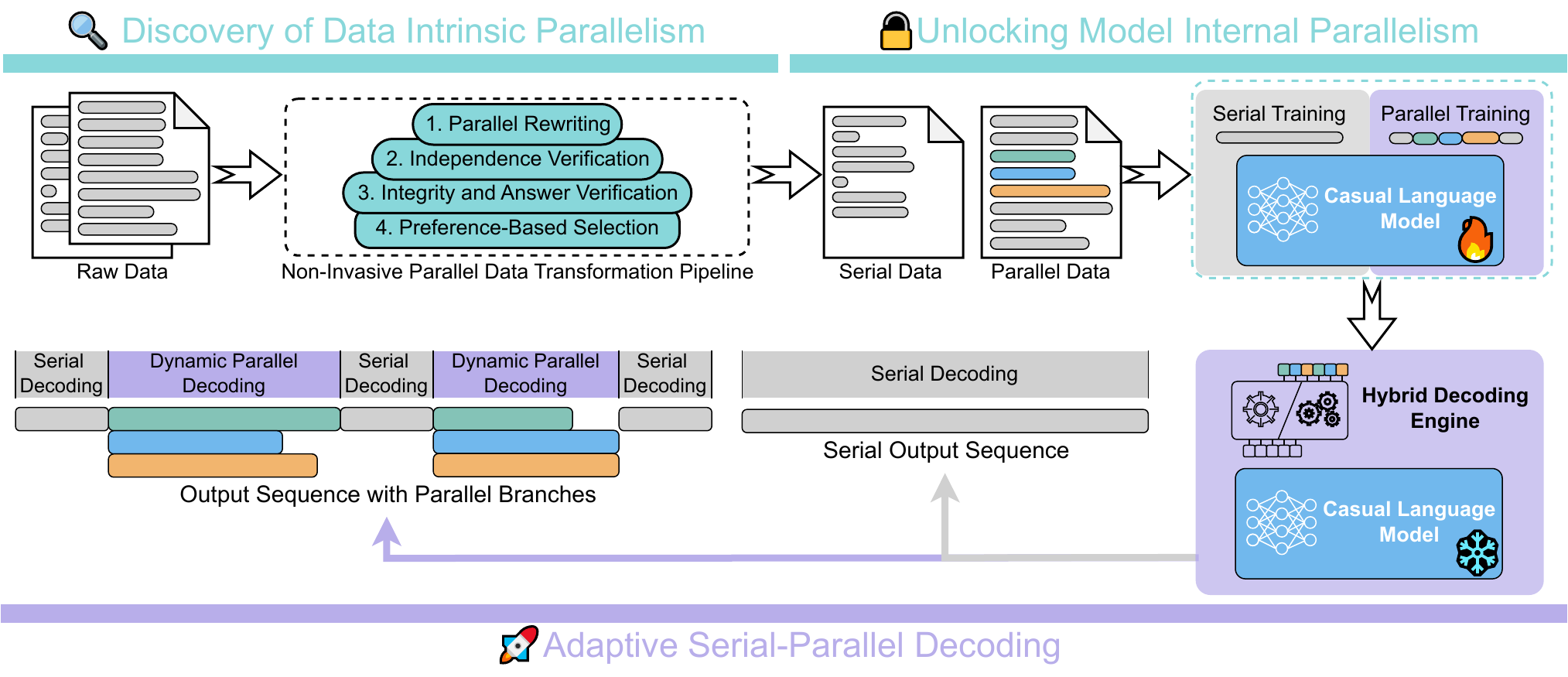}
    \caption{An Overview of Our \textbf{Adaptive Serial-Parallel Decoding} Framework.}
    \label{fig:methods_onverview}
\end{figure}
Our framework unlocks the intrinsic parallel capabilities of auto-regressive (AR) large language models through two key aspects: 
mining inherent parallelism in AR model responses and adapting internal model architectures. 
As shown in Figure \ref{fig:methods_onverview}, we achieve these objectives through  three key components below: 
\begin{enumerate}
\item \textbf{Non-invasive parallel data transformation pipeline. } 
This pipeline converts serial outputs into parallelizable corpora while preserving semantic integrity. 
Such curated training data significantly mitigates distribution shift relative to next-token prediction pretraining objectives, 
accelerating convergence and enhancing final performance.
\item \textbf{Internal Parallelization with Adaptive Serial-Parallel Decoding.} 
We introduce architectural modifications that ensure parallel decoding phases maintain behavioral consistency with native auto-regressive decoding.
This design principle maximizes utilization of capabilities acquired during pretraining and alignment stages without compromising generation quality.
\item \textbf{Hybrid Decoding Engine. } 
To enable efficient iterative transitions between serial and parallel modes, we propose a novel Hybrid Decoding Engine.
\end{enumerate}

\subsection{Natural Parallelism in AR LLM Responses}
\label{sec:para_data_pipe}
\begin{figure}[htbp]
    \centering
    \includegraphics[width=\textwidth]{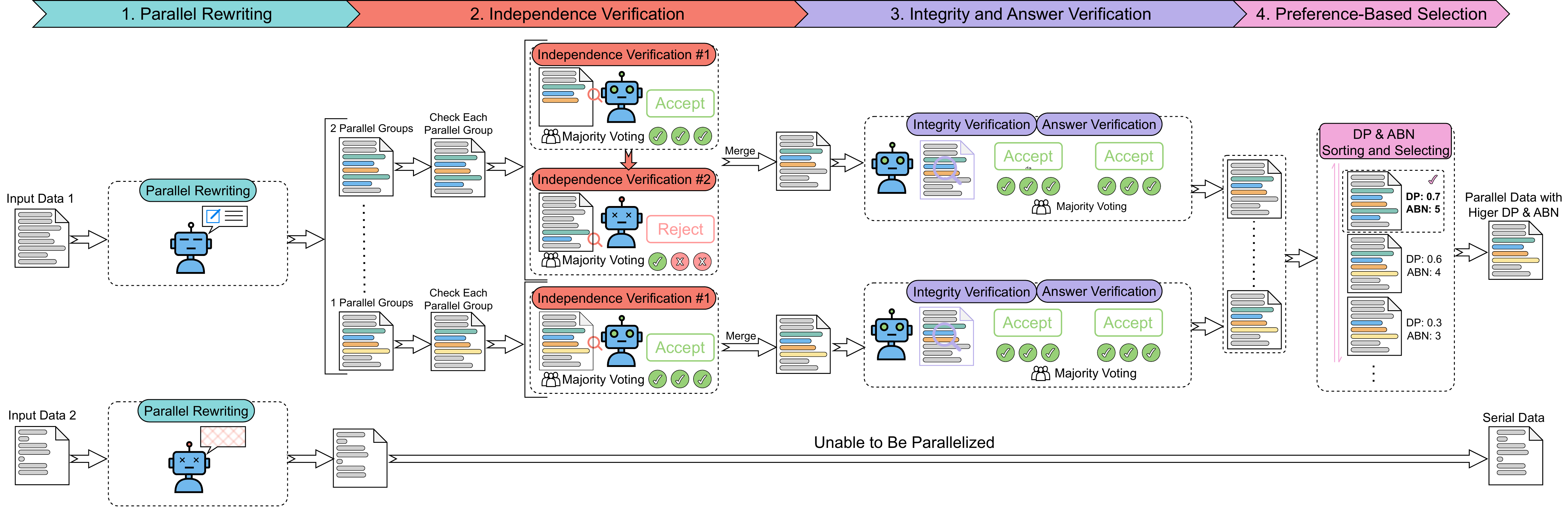}
    \caption{\textbf{Non-Invasive Parallel Data Transformation Pipeline.} }
    \label{fig:data_ppl}
\end{figure}

Despite the sequential nature of autoregressive language models, our analysis reveals significant inherent parallelism within their response generation process. 
As illustrated in Figure \ref{fig:ourmethod},  substantial portions of responses across diverse scenarios exhibit natural parallel structures that can be leveraged for concurrent generation.
To effectively exploit this latent parallelism, we propose a novel non-invasive parallelization framework that addresses three fundamental challenges:

\begin{enumerate}
    \item \textbf{Structural Decomposition. } 
    Identifying semantically independent components suitable for parallel generation while preserving logical dependencies.
    \item \textbf{Branch Independence. } 
    Maintaining mutual invisibility between parallel branches to ensure coherent and non-redundant generation.
    \item \textbf{Semantic Completeness. } 
    Guaranteeing comprehensive coverage of the original query intent across parallel branches. 
\end{enumerate}

Figure \ref{fig:data_ppl} depicts the four foundational stages of our proposed data curation pipeline.
\begin{enumerate}
\item \textbf{Parallel Rewriting. } Convert each data sample into serial-parallel groups. Multiple response samples are generated for each input. Samples with zero parallel groups are categoried as serial data; 
\item \textbf{Independence Verification. } Validate mutual independence within each parallel group. Parallel groups failing verification degenerate to serial form. 
Samples with all parallel groups failing degenerate to serial data;
\item \textbf{Integrity and Answer Verification. }Verify structural completeness of parallel groups and perform answer consistency checking;
\item \textbf{Preference-based Selection. }
Final data selection retains highest-ranked samples based on Degree of parallelization(DP) and Average Branch Num (ABN).
\end{enumerate}
This comprehensive pipeline systematically mines and validates the intrinsic parallelizable patterns within model responses through multi-stage verification, ensuring both semantic independence and structural completeness. To reduce bias in model judgment, we sample three output paths for each validation result. We then use the majority voting method to decide whether the validation is accepted. The validation is considered successful only when all sampled output paths unanimously agree. By leveraging these inherent parallel structures, our approach facilitates the construction of high-quality parallel training data while preserving the model's native capabilities and response characteristics.
\subsection{Internal Parallel With Adaptive Serial-Parallel Decoding}

\begin{figure}[htb]
    \centering
    \begin{subfigure}{0.66\textwidth}
        \includegraphics[width=\linewidth]{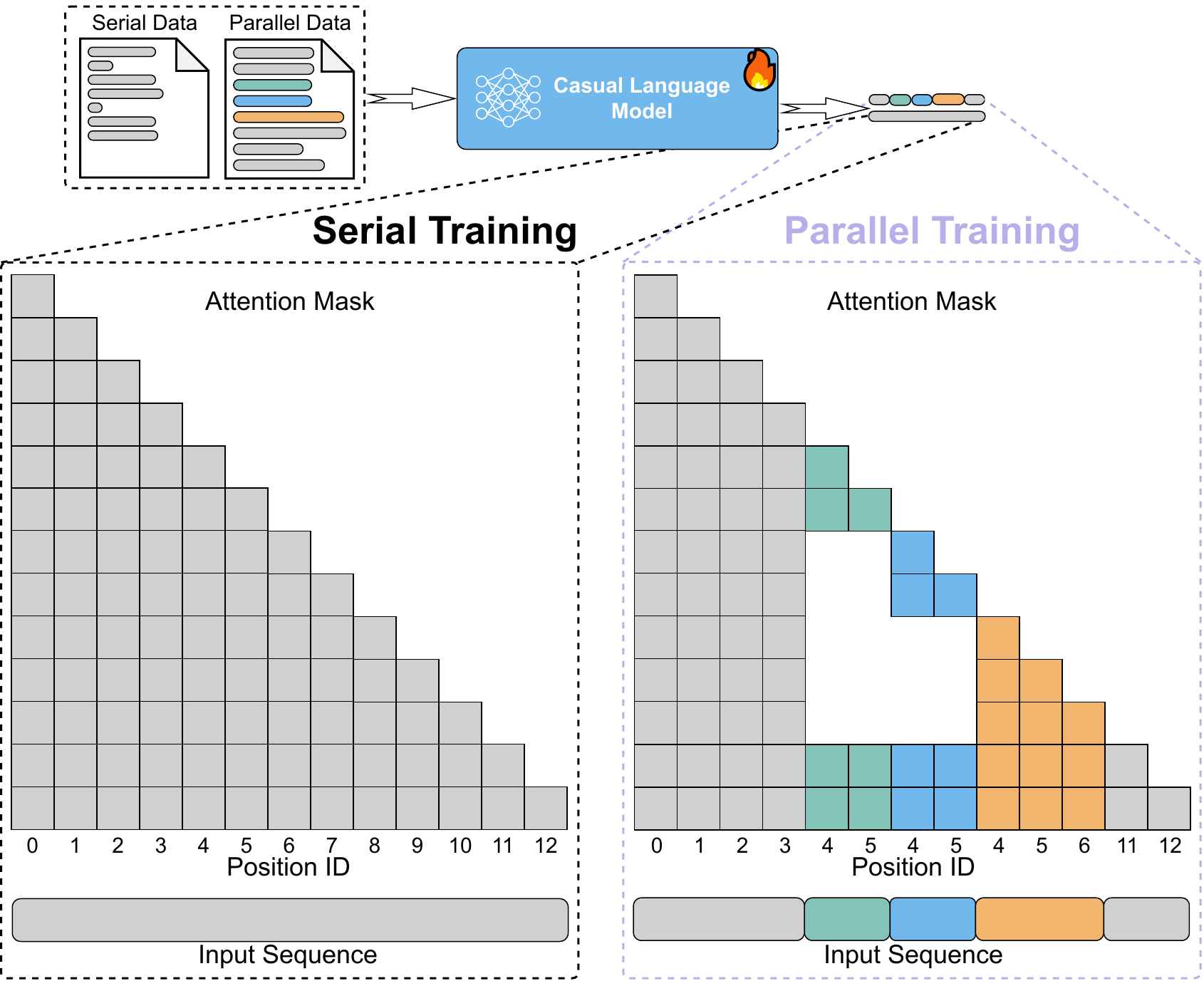}
        \caption{Training Details of Processing Sequential and Parallel Data.}
        \label{fig:train_detials}
    \end{subfigure}
    \hfill
    \begin{subfigure}{0.33\textwidth}
        \includegraphics[width=\linewidth]{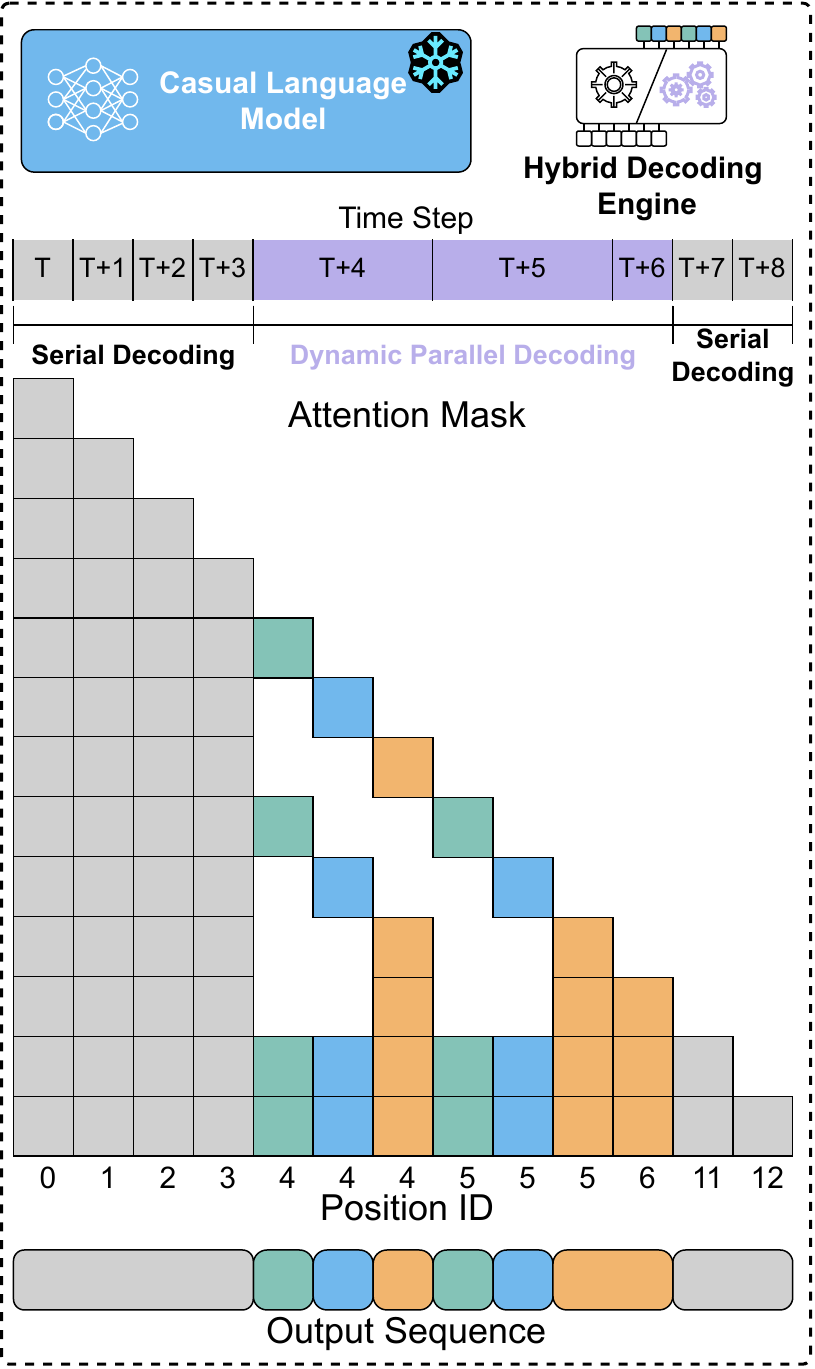}
        \caption{Hybrid Decoding Engine.}
        \label{fig:test_details}
    \end{subfigure}
    \caption{The Training Process and Our \textbf{Hybrid Decoding Engine} in Inference.\quad
      \textbf{\textit{Training phase. }} Tokens within each parallel branch are arranged sequentially due to teacher forcing;\quad
      \textbf{\textit{Inference phase. }} The autoregressive nature within each branch and parallelism between branches making multiple tokens generated simultaneously, 
      resulting in interleaved parallel token arrangements.}
    \label{fig:model_arc}
\end{figure}

\textbf{Model Architecture for Native Parallelization. }
\label{sec:model_arc}
After obtaining the parallelized corpus described in Section \ref{sec:para_data_pipe}, we need to modify the model architecture to enable efficient serial-parallel decoding.
As shown in Figure \ref{fig:model_arc}, we need to handle both the visibility between branches and the position encoding within it.
To better leverage the capabilities of the native autoregressive model during parallel phases while supporting seamless transitions between parallel and sequential modes,
we introduce two key components:
(1) an internal parallel mask for branch-independent parallel decoding, and
(2) shared positional encodings across parallel branches at the same timestamp.

\textbf{Preliminaries. }
The generation process comprises a sequence of interleaved \textbf{stages}, each decoding in serial or parallel mode.
In serial stages, generation proceeds through a single \textbf{main branch} in an autoregressive manner.
During parallel stages, the model simultaneously decodes multiple \textbf{parallel branches}, enabling concurrent token generation across different aspects.
Here, $b(i)$ denotes the branch index of token $i$ and $t$ denotes temporal timestamps, 
$stage\_start(i)$ denotes the starting position of the stage where token $i$ is located,
 and $\Delta i$ represents the relative position of token $i$ within its stage.
$P_t$ represents the number of tokens being decoded simultaneously at time $t$. 
\begin{align}
\text{Attn}(Q, K, V) &= \text{softmax}\left( \frac{QK^\top}{\sqrt{d_k}} + M \right) V \label{eq:attn} \\
M_{i,j} &= \begin{cases}
0 & \text{if } \mathcal{S}(b(i), b(j)) = 1 \ \text{and} \ \text{pos}(i) > \text{pos}(j) \\
-\infty & \text{otherwise}
\end{cases} \label{eq:mask} \\
\mathcal{S}(b(i), b(j)) &= \begin{cases}
1 & \text{if } b(i) \text{ in main branch}  \\
1 & <b(i), b(j)> \ \text{in same parallel branch} \\
1 & <b(i), b(j)> \ \text{in different stage} \\
0 & \text{otherwise}
\end{cases} \label{eq:visibility} \\
\text{pos}(i) &= \begin{cases}
\sum_{t=1}^{i-1} P_t + 1 & \text{if i in main branch} \\ 
stage\_start(i) + \Delta i & \text{if i in para branch} \\
\end{cases} \label{eq:position}
\end{align}

Equations \ref{eq:attn}-\ref{eq:position} formally define our parallel attention and shared position mechanism. 
We extend the vanilla causal attention by incorporating a visibility function $\mathcal{S}$.
As shown in Eq. \ref{eq:mask}, $M_{i,j}$  is only visible when the visibility function $\mathcal{S}$ equals 1 and position j precedes position i.
Eq. \ref{eq:visibility} defines the visibility between tokens, where the main branch can see all other branches, 
while for parallel branches, visibility is enabled when $<b(i), b(j)>$ spans across stages or belongs to the same parallel branch.
For position encoding, as shown in Eq. \ref{eq:position}, tokens in the main branch maintain absolute positions in the flattened sequence, 
while parallel branches synchronize their position encodings at each timestamp. More intuitively, we can interpret Equations \ref{eq:attn}-\ref{eq:position} from the following two aspects.

\textbf{The Parallel Branch's Perspective. }
Eq. \ref{eq:mask} - \ref{eq:visibility} ensures mutual invisibility across branches during concurrent generation, 
while Eq. \ref{eq:position} maintains monotonically increasing position IDs with shared position encodings at each timestamp.
These mechanisms preserve sequential decoding patterns within each branch, keeping the decoding behavior consistent with native autoregressive generation.

\textbf{The Main Branch's Perspective. }
After completing all parallel branches, the model seamlessly transitions to standard autoregressive decoding mode,
allowing main branch tokens to attend to the complete history of preceding tokens.
This dual-nature design enables the main branch to maintain its native autoregressive causality while
leveraging shared positional encodings across parallel branches to incorporate additional contextual information.

\subsection{Hybrid Decoding Engine}

Having established the foundational data processing pipeline and model architectural modifications for internal parallel decoding, we designed a hybrid decoding engine that enables seamless and adaptive transitions between serial and parallel processing modes. To enable parallelization, we augment the vocabulary with six special tokens: \texttt{<Title>}, \texttt{</Title>}, \texttt{<Branch>}, \texttt{</Branch>}, \texttt{<Para>}, and \texttt{</Para>}.

During inference, when the model determines that the current response can be parallelized, it generates several parallel titles, each enclosed by \texttt{<Title>} and \texttt{</Title>}. After all branch titles are generated, the model outputs a \texttt{<Para>} token, turning the engine into parallel decoding mode. For each parallel branch $i$, we set the prefix "\texttt{<Branch>}T$i$: " to help the model identify which branch to generate, where T$i$ is the corresponding title for branch $i$. A branch is considered complete when the \texttt{</Branch>} token is reached. Once all branches are finished, a \texttt{</Para>} token is appended to the output sequence, switching the engine back to serial decoding mode.

The hybrid engine implements the branch-independent parallel mask and synchronized position-id mechanism proposed in Section \ref{sec:model_arc} to support parallel decoding. Each parallel token attends only to tokens from both the main branch and its respective branch, with sequentially increasing position-ids. This allows each branch to maintain the same generation pattern as native autoregressive models from its own perspective. After all parallel branches complete decoding, sequential decoding resumes on the main branch, where the position ID of the first token reflects its actual position in the complete sequence. Our hybrid decoding engine supports iterative serial-parallel switching for optimal efficiency.

\section{Experiments}



\begin{figure}[htbp]
    \centering
    \includegraphics[width=0.96\textwidth]{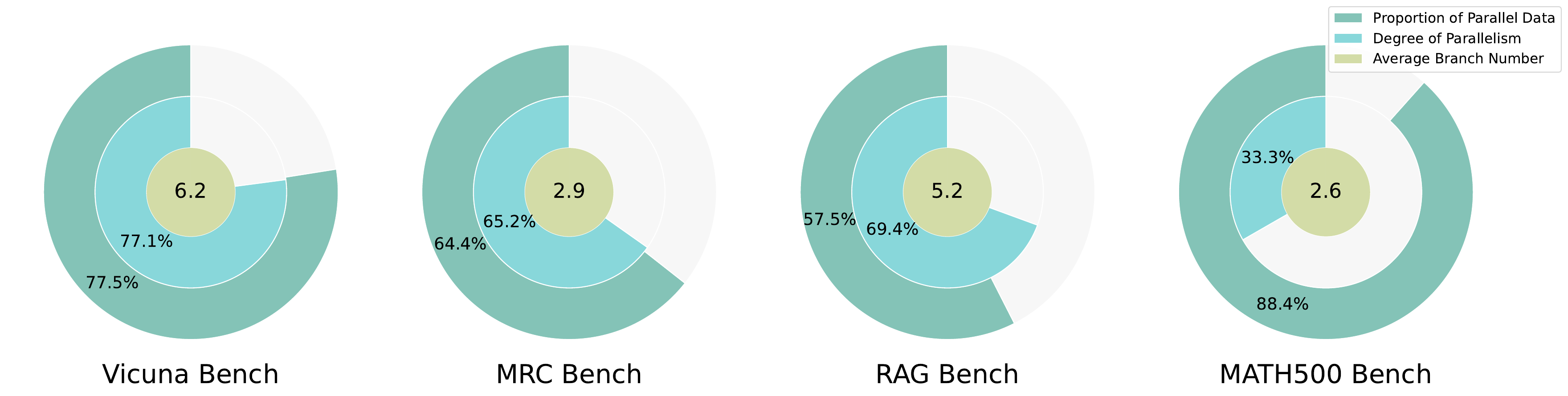}
    \caption{Parallelization Patterns in Our Trained Parallel Model's Responses Across Different               Scenarios}
    \label{fig:data_analys_model}
\end{figure}

Building on our analysis of parallelization potential (Figure \ref{fig:ourmethod}),
our method effectively leverages intrinsic parallelism while maintaining generalizability across diverse scenarios (Figure \ref{fig:data_analys_model}).
Results on Vicuna\citep{chiang2023vicuna}, our Internal MRC, and RAG\protect\footnotemark[1] benchmark demonstrate optimal parallelism utilization.
The Mathematical Reasoning (MATH500)\citep{hendrycks2021measuring} benchmark exhibits the highest parallelization constraints among evaluated tasks,
with 88.4\% PPD (+4.2\%), 33.3\% DP (+2.7\%), and 2.6 ABN (-0.1),
which exhibits strong alignment with the statistical patterns shown in Figure \ref{fig:ourmethod}. 
These quantitative metrics provide compelling evidence that \textbf{our approach successfully harnesses 
the inherent parallelization capabilities of LLMs while preserving the fundamental characteristics of native autoregressive generation.}

\subsection{Experimental Setup}
\textbf{Dataset. }
Following \citep{liu2024APAR}, we use ShareGPT Vicuna dataset\protect\footnotemark[1]  as our training data. 
This dataset encompasses instructions across various scenarios including STEM, roleplay, reasoning, and extraction tasks. 
Additionally, we utilize our internal Machine Reading Comprehension (MRC) dataset for multi-dimensional validation. 
All data processing is conducted using our proposed non-invasive parallelization transformation pipeline to obtain parallelized data. 
For fair comparison, we create corresponding sequential data by removing special parallel tokens.
\footnotetext[1]{\url{https://huggingface.co/datasets/anon8231489123/ShareGPT_Vicuna_unfiltered}}
\footnotetext[2]{\url{https://huggingface.co/datasets/neural-bridge/rag-dataset-12000}}

\textbf{Evaluation. }
We conduct comprehensive evaluations using established benchmarks including Vicuna Bench \citep{chiang2023vicuna} and MT Bench \citep{zheng2023judging}, following the evaluation protocol of \citep{liu2024APAR}. 
Furthermore, to evaluate models' generalization capability , 
we introduce an out-of-domain benchmark, namely the RAG Bench. 
This benchmark consists of the first 200 questions with corresponding context sampled from rag-dataset-12000\protect\footnotemark[2].
Additionally, we evaluate on Internal Logical Reasoning (LR) and AI Search benchmarks.
Our evaluation framework encompasses two primary dimensions: model effectiveness and computational efficiency.
\begin{itemize}
\item \textbf{Effectiveness Metrics.  }
For public benchmarks, we utilize the LLM-as-judge evaluation framework proposed by \citep{zheng2023judging} to quantify response quality, maintaining methodological consistency with \citep{liu2024APAR}.
All evaluations are conducted using Qwen3-235B-A22B \citep{yang2025qwen3} to ensure balanced assessment.
For Internal benchmarks, we implement human evaluation protocols to mitigate potential model-based assessment biases.
\item \textbf{Efficiency Metrics. }
We employ \textbf{T}okens-\textbf{P}er-\textbf{S}econd (\textbf{TPS}) as the primary throughput metric.
For parallel models, we further introduce four additional metrics to characterize parallelization performance:
\textbf{P}arallel-\textbf{T}okens-\textbf{P}er-\textbf{S}econd (\textbf{P-TPS}, TPS in parallel stage),
\textbf{D}egree of \textbf{P}arallelism (\textbf{DP}, ratio of parallel to total tokens),
\textbf{A}verage \textbf{B}ranch \textbf{N}um (\textbf{ABN}, average parallel branch num),
\textbf{P}roportion of \textbf{P}arallel \textbf{D}ata (\textbf{PPD}).
\end{itemize}

\begin{table}[htb]
  \setlength{\tabcolsep}{4pt}
  \centering
  \caption{Performance Comparison on MT Bench and Vicuna Bench. 
  The  \textbf{bold} and \underline{underlined} values denote the best and second-best results, respectively. 
  \textbf{SoT} represents evaluation results using official code on corresponding benchmarks, 
  \textbf{V-APAR} shows results using open-source parameters,
   and \textbf{V-APAR$^{}$*} corresponds to our fine-tuned APAR model using query-aligned data augmented by Qwen3-235B-A22B \citep{yang2025qwen3}.
    \textbf{Ori} denotes the original model, while \textbf{Seq} refers to sequential fine-tuning without parallel tokens.
     Models evaluated include Vicuna-1.3-7B (V) and Qwen2.5-7B-Instruct (Q). CS is short for Common-Sense on Vicuna Bench.}
    \begin{tabular}{c|cccccc|ccc}
    \toprule
    \multicolumn{10}{c}{\bf MT Bench} \\
    \midrule
    \diagbox{\scriptsize \bf Task}{\scriptsize \bf Model} &  V-Ori &  V-Seq &  V-APAR &  SoT &  V-APAR* & \bf V-ASPD &  Q-Ori &  Q-Seq & \bf Q-ASPD \\
    \midrule
    \bf Coding & \textbf{3.50} & \underline{3.10} & 2.70  & 3.00  & 2.55  & 2.90  & \textbf{7.15}  & 6.60  & \underline{6.70}  \\
    \bf Extraction & \underline{5.10} & 4.40 & \textbf{5.20}  & 3.08  & 5.05  & 4.80  & \underline{7.45}  & 6.93  & \textbf{7.80}  \\
    \bf Humanities & 6.55 & \textbf{8.40} & 6.40  & 6.30  & \underline{8.10}  & 7.95  & 8.70  & \textbf{9.15}  & \underline{9.05}  \\
    \bf Math  & 2.95 & \textbf{3.05} & 2.55  & 2.75  & 2.50  & \underline{3.00}  & \underline{8.50}   & 8.45 & \textbf{8.90}  \\
    \bf Reasoning & 4.65 & 4.80 & \underline{5.45}  & 4.50  & 4.75  & \textbf{5.85}  & 6.85  & \textbf{7.25}  & \underline{7.05}  \\
    \bf Roleplay & 5.30 & \underline{7.10} & 5.60  & 5.60  & 6.80  & \textbf{7.25}  & 8.10  & \underline{8.40}  & \textbf{8.48}  \\
    \bf Stem  & 5.55 & 6.20 & 5.55  & 5.85  & \textbf{6.50}  & \underline{6.30}  & 7.80  & \textbf{8.60}  & \underline{8.50}  \\
    \bf Writing & 5.30 & \textbf{7.65} & 5.60  & 4.75  & \underline{6.80}  & 6.70  & 8.00  & \underline{8.50}  & \textbf{8.75}  \\
    \midrule
    \bf Mean  & 4.86 & \textbf{5.59} & 4.88  & 4.48  & \underline{5.38}  & \textbf{5.59}  & 7.82  & \underline{7.98}  & \textbf{8.15}  \\
    \midrule
    \multicolumn{10}{c}{\bf Vicuna Bench} \\
    \midrule
    \diagbox{\scriptsize \bf Task}{\scriptsize \bf Model} &  V-Ori &  V-Seq &  V-APAR &  SoT &  V-APAR* & \bf V-ASPD &  Q-Ori &  Q-Seq & \bf Q-ASPD \\
    \midrule
    \bf Coding & \textbf{4.29} & 3.71 & \underline{4.14}  & 3.43  & 3.71  & 3.71  & \textbf{9.00}  & 8.29  & \underline{8.71}  \\
    \bf CS    & 7.40 & \textbf{9.00} & 7.20  & 7.30  & \underline{8.90}  & \underline{8.90}   & \underline{9.00}  & \textbf{9.30}  & \underline{9.00}  \\
    \bf Counterfactual & 5.10 & \underline{8.50} & 5.30  & 5.30  & 8.40  & \textbf{8.60}   & \underline{8.40}  & \textbf{9.10}  & \textbf{9.10}  \\
    \bf Fermi & \textbf{5.50} & 4.50 & 4.60  & 4.40  & 5.00  & \underline{5.20}   & 7.70  & \textbf{8.20}  & \underline{8.10}  \\
    \bf Generic & 7.30 & \textbf{9.20} & 7.40  & 7.40  & \underline{8.95}  & 8.80  & 8.70  & \textbf{9.55}  & \underline{9.40}  \\
    \bf Knowledge & 7.30 & \textbf{9.10} & 7.20  & 7.50  & 8.60  & \underline{9.00}  & 9.00  & \textbf{9.40}  & \underline{9.10}  \\
    \bf Math  & \underline{2.67} & 2.33 & \textbf{3.33}  & 2.00  & 2.33  & \underline{2.67}  & \underline{9.67}  & \textbf{10.00}  & \textbf{10.00}  \\
    \bf Roleplay & 7.20 & \textbf{9.20} & 6.60  & 6.70  & \underline{9.00}  & \underline{9.00}  & 9.00  & \underline{9.30}  & \textbf{9.40}  \\
    \bf Writing & 6.10 & \underline{8.80} & 6.60  & 5.80  & \underline{8.80}  & \textbf{9.00}  & 8.20  & \textbf{9.20}  & \underline{9.00}  \\
    \midrule
    \bf Mean  & 6.21 & \underline{7.70} & 6.10  & 5.93  & 7.62  & \textbf{7.74}  & 8.65  & \textbf{9.11}  & \underline{9.03}  \\
    \bottomrule
    \end{tabular}%
  \label{tab:bench_scores_all}%
\end{table}%
\textbf{Models. }
Following \citep{liu2024APAR}, we employ Vicuna-V1.3-7B \citep{zheng2023judging} as our base model to ensure fair comparison across different approaches. 
To demonstrate the cross-architecture generalization capability of our method, we further conduct experiments using Qwen2.5-7b-Instruct \citep{team2024qwen2} as an additional foundation model.

\textbf{Implementation Details. }
During the training phase, we employed an initial learning rate of 1e-5 and a global batch size of 16 for three training epochs.
The training process utilized cosine learning rate scheduling with a warmup ratio of $0.1$.
The context length for training data was set to 8k tokens.
\textbf{Both sequential and parallel models used identical parameter configurations to ensure fair comparison.}
For inference, we consistently set the temperature parameter to 0.7, top\_K to 20, and top\_P to 0.8 across all experiments.

\subsection{Main Results}

Tables \ref{tab:bench_scores_all} present a systematic evaluation across model architectures and methodologies.
We evaluate our approach using two base models: Vicuna-1.3-7B (V) and Qwen2.5-7B-Instruct (Q). 
For each base model, we compare the original model (Ori), sequential fine-tuned model (Seq), SoT\citep{ning2023skeleton}, APAR\citep{liu2024APAR}. 
Additionally, to ensure fair comparison, we utilize APAR's official codebase and enhance its training data quality using Qwen3-235B-22A to obtain APAR*.
The experimental results demonstrate the following:

\begin{figure}[!htp]
    \centering
    \begin{subfigure}[b]{0.94\textwidth}
        \centering
        \includegraphics[width=\textwidth]{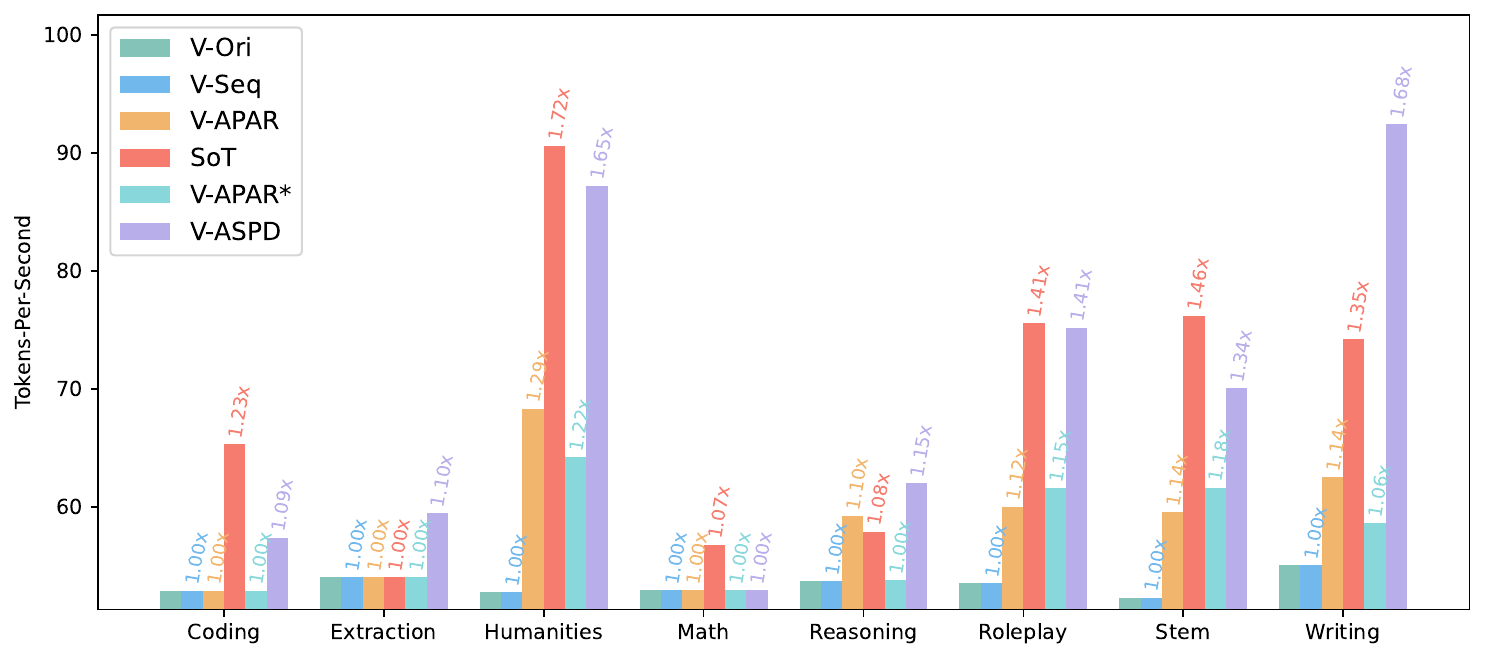}
        \vspace{-2em}
        \caption{MT Bench.}
        \label{fig:all_speed_mt}
    \end{subfigure}
    \vspace{-1em}
    \begin{subfigure}[b]{0.94\textwidth}
        \centering
        \includegraphics[width=\textwidth]{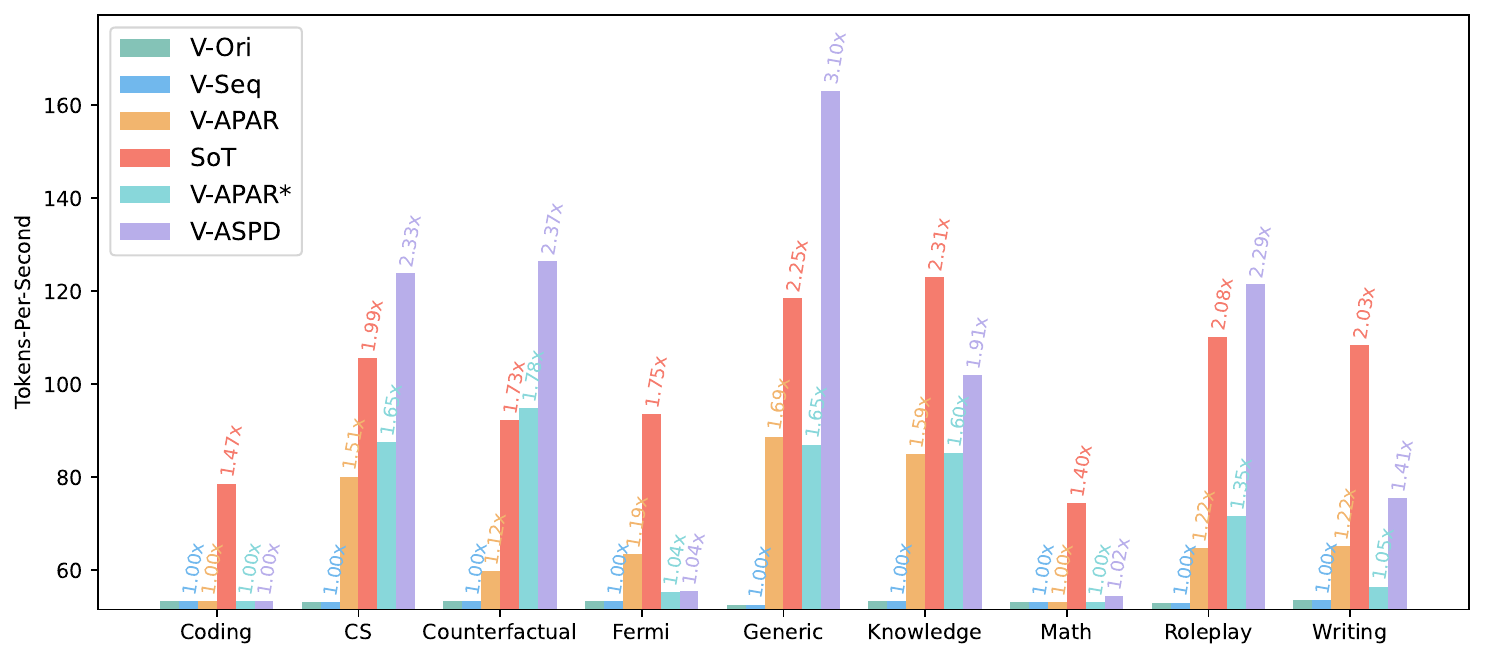}
        \vspace{-2em}
        \caption{Vicuna Bench.}
        \label{fig:all_speed_vi}
    \end{subfigure}
    \caption{\textbf{Token Speed Analysis on Different Benchmarks. } Our method achieves the best speedup performance on MT Bench and Vicuna Bench. 
    For specific tasks, the speedup ratios reach 1.68-3.10x in scenarios like Generic, Writing, and Common-Sense (CS).}
  \label{fig:token_speed_analysis}
\end{figure}

\begin{figure}[htbp]
    \centering
    \begin{subfigure}{0.30\textwidth}
        \includegraphics[width=\linewidth]{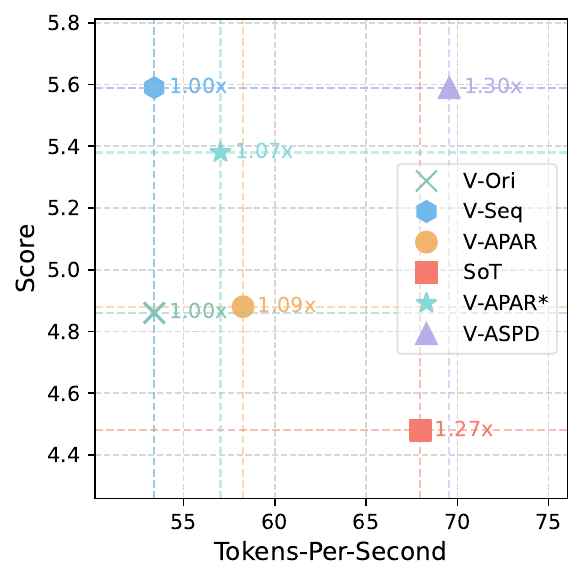}
        \caption{MT Bench.}
        \label{fig:mean_speed_vi}
    \end{subfigure}
    \hfill
    \begin{subfigure}{0.30\textwidth}
        \includegraphics[width=\linewidth]{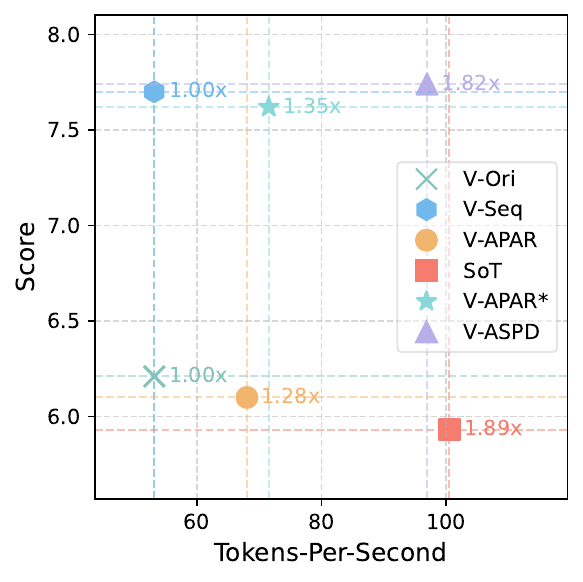}
        \caption{Vicuna Bench.}
        \label{fig:mean_speed_mt}
    \end{subfigure}
    \hfill
    \begin{subfigure}{0.30\textwidth}
        \includegraphics[width=\linewidth]{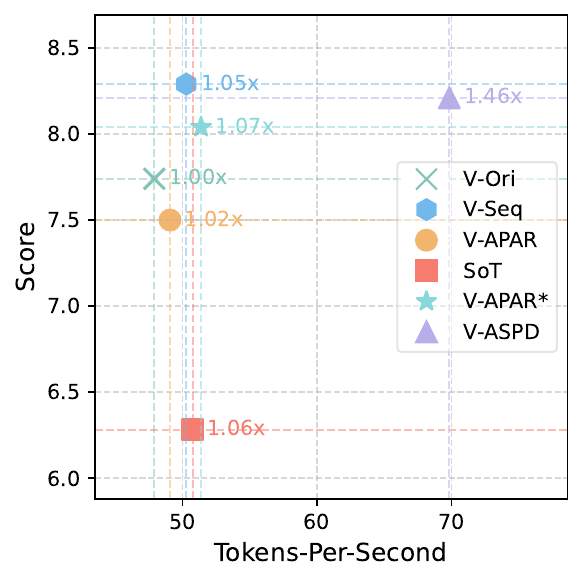}
        \caption{RAG Bench.}
        \label{fig:mean_speed_rag}
    \end{subfigure}
    \vspace{-1em}
    \caption{Speed-Quality Trade-off Analysis Across Different Parallel Decoding Methods On Various Benchmarks.
         Our method achieves the fastest average speed while maintaining better accuracy than other parallelization approaches, nearly matching the accuracy of sequential models fine-tuned on the same data.}
    \label{fig:mean_speed}
\end{figure}

\begin{itemize}
\item \textbf{Token Speed Analysis on Different Benchmarks. } 
As shown in Figure \ref{fig:token_speed_analysis}, 
our method achieves significant speedup across different subtasks in both MT Bench and Vicuna Bench evaluations. 
Specifically, compared to the native sequential model, 
we observe speedup ratios ranging from 1.0x to 3.10x across Coding to Writing, outperforming APAR, APAR* and SoT on most subtasks.

\item \textbf{Superior Response Quality Compared to Previous Approaches. } Both our fine-tuned parallel and serial models outperform the original model,
 establishing a solid foundation for our subsequent evaluations. Our parallel model demonstrates superior performance compared to the APAR and SoT method, 
achieving a 14.55\% and 24.78\% improvement on the MT Bench, a 26.89\% and 30.52\% enhancement on the Vicuna Bench. 
highlighting the effectiveness of our approach.


\item \textbf{Superior Acceleration While Maintaining Quality. }
As shown in Figure \ref{fig:mean_speed}, ASPD achieves superior speed-quality balance across all benchmarks, with average acceleration ratios of 1.30x to 1.82x. 
Specifically,  On Vicuna Bench, ASPD reaches 1.82x speedup, surpassing APAR (1.28x) and APAR* (1.35x), while nearly matching SoT's 1.89x that requires intensive memory usage. 
However, SoT's quality score of 5.93 is notably lower than both ASPD (7.74) and baseline (6.21), limited by its rigid prompt design.
 On RAG Bench, SoT's speedup drops to 1.06x compared to ASPD's 1.46x, due to its inefficient two-stage process requiring repeated prefilling. These results demonstrate ASPD's consistent advantages in balancing efficiency and quality.
The consistent performance gains across diverse evaluation scenarios highlight the robustness and effectiveness of our parallel decoding framework.
 
 \item \textbf{Cross-Domain Generalization. }
As shown in Figure \ref{fig:mean_speed_rag}, we observed that other competing methods face generalization issues on the out-of-domain RAG Bench, such as declining scores or acceleration ratios. In contrast, our method not only achieves comparable generation quality to V-Seq but also maintain the highest speedup. This demonstrates our method's strong out-of-domain generalization capability, suggesting its potential for broader application.

\item \textbf{Cross-Model Evaluation. }
The results in Table \ref{tab:bench_scores_all}, based on Qwen2.5-7B-Instruct (Q) model, demonstrate ASPD's consistent efficacy across diverse architectures. On MT Bench, Q-ASPD scores 8.15, outperforming both Q-Ori and Q-Seq with a 2.1\% improvement over Q-Seq. Vicuna Bench shows comparable performance between Q-ASPD (9.03) and Q-Seq (9.11), with a mere 0.9\% difference, confirming ASPD's output quality preservation. The minimal performance variation across benchmarks underscores ASPD's robustness.

\end{itemize}

Interestingly, when we updated APAR's training data using the higher-quality response of Qwen3-235B-22A, 
its performance scores improved by 10.25\% compared to the original APAR,
yet its speedup ratio remained largely unchanged and lower than ASPD (1.07x vs 1.30x).
Analysis of responses on MT Bench revealed that APAR* achieved a parallel response ratio of 24.75\%,
while ours reached 59.38\%, explaining the difference in speedup ratios.
This demonstrates that merely using higher quality data, 
without further exploring intrinsic parallelism and adapting corresponding model architectures, 
makes it difficult to achieve significant speedup improvements.


\subsection{Generalization Across Different Benchmarks}
To evaluate the generalization capability of parallel processing across different benchmarks, 
we conducted training on our internal MRC benchmark and tested generalization performance on MRC, LR, and AI Search benchmarks, using Qwen2.5-7B-Instruct as our base model.
Specifically, MRC is designed to evaluate a wide range of reading comprehension skills,
including various question types such as single passage understanding, multi-information extraction, refusal to answer, etc.
Logical Reasoning (LR) contains problems requiring step-by-step deductive reasoning, 
while AI Search resembles RAG scenarios where responses are generated by analyzing multi-source search results, 
typically presented in an itemized format.

As shown in Table \ref{tab:internal_scores}, on the homogeneous MRC benchmark,
 ASPD achieves a 1.35x speedup in TPS compared to the original model, 
while maintaining performance within 1\% of the fine-tuned sequential model. 
Cross-validation using LR and AI Search benchmarks demonstrates acceleration ratios of 1.15x and 1.45x respectively.

\footnotetext[1]{\url{https://huggingface.co/datasets/zwhe99/amc23}}
\footnotetext[2]{\url{https://huggingface.co/datasets/HuggingFaceH4/aime_2024}}
\footnotetext[3]{\url{https://huggingface.co/datasets/yentinglin/aime_2025}} 
\footnotetext[4]{\url{https://huggingface.co/datasets/open-r1/OpenR1-Math-220k} }

For the AI Search test set, which contains more questions requiring itemized responses, we observe increased opportunities for parallel output generation. This leads to improved parallelization metrics including PPD, DP, and ABN (3.21 in AI Search versus 2.40 in Logical Reasoning), as illustrated in Table \ref{tab:internal_para_metrics}. Consequently, both TPS and P-TPS metrics show significant improvements. These results indicate that the effectiveness of our parallelization approach varies depending on the specific task characteristics.
\begin{table}[htbp]
  \begin{minipage}{0.49\textwidth}
    \setlength{\tabcolsep}{3pt}
    \centering
    \caption{Comprehensive Parallelization Metrics on \newline \quad 
              Our Internal Test Sets}
    \begin{tabular}{c|cccc}
    \toprule
    \bf Test Set & \bf PPD    & \bf DP    & \bf ABN   & \bf P-TPS \\
    \midrule
    MRC   & 64.40  & 65.21  & 2.86  & 109.32\textsubscript{2.18x} \\
    LR & 44.59  & 55.80  & 2.40  & 101.72\textsubscript{1.92x} \\
    AI Search & 64.50  & 67.69  & 3.21  & 97.27\textsubscript{2.41x} \\
    \bottomrule
    \end{tabular}%
    \label{tab:internal_para_metrics}%
  \end{minipage}%
  \begin{minipage}{0.49\textwidth}
    \setlength{\tabcolsep}{3pt}
    \centering
    \caption{Human Evaluation on Our Internal Test Sets}
    \begin{tabular}{c|cc|cc|cc}
    \toprule
    \multirow{2}[2]{*}{\bf Model} & \multicolumn{2}{c|}{\bf MRC} & \multicolumn{2}{c|}{\bf LR} & \multicolumn{2}{c}{\bf AI Search} \\
          &  ACC &  TPS   &  ACC &  TPS   &  Score &  TPS \\
    \midrule
    Q-Ori & 64.40  & 50.18  & 53.50  & 53.06  & 7.85  & \underline{40.29}  \\
    Q-Seq & \textbf{73.20}  & \underline{50.53}  & \textbf{56.69}  & \underline{54.32}  & \textbf{8.23}  & 39.89  \\
    \bf Q-ASPD & \underline{72.40}  & \textbf{67.91} & \underline{55.41} & \textbf{61.26} & \underline{8.15}  & \textbf{58.56} \\
    \bottomrule
    \end{tabular}%
    \label{tab:internal_scores}%
  \end{minipage}
\end{table}%

\subsection{Parallelism at the Reasoning Frontier. }
While previous approaches like APAR excluded mathematical and coding tasks from parallel data processing, 
we observe that our concurrent research work Multiverse\citep{yang2025multiverse} has also directed attention toward mathematical
problem parallelization. 
To thoroughly investigate this domain, we extended our model architecture to Qwen2.5-32B-Instruct\citep{team2024qwen2} 
and conducted comprehensive validation on these complex reasoning domains.
Our evaluation includes a diverse range of benchmarks, including GPQA\citep{rein2024gpqa}, MATH500\citep{hendrycks2021measuring}, 
and competition-level mathematics such as AMC23\protect\footnotemark[1], AIME24\protect\footnotemark[2], and AIME25\protect\footnotemark[3].
For training data, we utilize OpenR1-Math-220K\protect\footnotemark[4] \citep{openr1},
which was generated by prompting DeepSeek-R1\citep{guo2025deepseek} with NuminaMath1.5 \citep{numina_math_datasets} as the query set. 
The model was trained for 9 epochs with a global batch size of 88, using a 12k context window and a learning rate of 1e-5.
\begin{table}[htbp]
  \begin{minipage}{0.30\textwidth}
    \setlength{\tabcolsep}{3.5pt} 
    \centering
    \caption{Math Bench Performance with Qwen2.5-32B-Intruct.}
    \vspace{0.5em} 
      \begin{tabular}{c|ccc}
      \toprule
      \textbf{\diagbox{\scriptsize \bf Bench}{\scriptsize \bf Model}} & Ori &  Seq & \bf ASPD \\
      \midrule
    \bf MATH500 & 82.00  & \textbf{94.40}  & \underline{94.00}  \\[0.3em]
    \bf AMC23 & 62.19  & \textbf{89.69}  & \underline{89.38}  \\[0.3em]
    \bf GPQA  & 48.99  & \underline{61.11}  & \textbf{65.66}  \\[0.3em]
    \bf AIME2024 & 17.50  & \underline{58.75}  & \textbf{62.08}  \\[0.3em]
    \bf AIME2025 & 12.50  & \underline{47.92}  & \textbf{50.00}  \\
      \bottomrule
      \end{tabular}%
    \label{tab:math_scores}%
  \end{minipage}
  \hspace{0.04\textwidth} 
  \begin{minipage}{0.48\textwidth}
    \centering
    \caption{Comparative Analysis of Parallelization Metrics on Mathematical Benchmarks between \textbf{ASPD} and \textbf{Seq.}}
    \vspace{0.5em} 
      \begin{tabular}{c|ccccc}
      \toprule
      \textbf{\diagbox{\scriptsize \bf Bench}{\scriptsize \bf Metric }} & {\bf PPD} & \textbf{\bf DP} & \textbf{\bf ABN} & \textbf{\bf TPS} & \textbf{\bf P-TPS} \\
      \midrule
    \bf MATH500 & 88.40  & 33.30  & 2.61  & 27.14\textsubscript{1.17x} & 43.03\textsubscript{1.86x} \\[0.3em]
    \bf AMC23 & 84.38  & 22.24  & 2.80  & 21.93\textsubscript{1.11x} & 39.30\textsubscript{1.99x} \\[0.3em]
    \bf GPQA  & 66.16  & 32.70  & 2.88  & 22.06\textsubscript{1.13x} & 36.57\textsubscript{1.88x} \\[0.3em]
    \bf AIME2024 & 65.42  & 8.84  & 2.48  & 16.43\textsubscript{1.04x} & 24.37\textsubscript{1.54x} \\[0.3em]
    \bf AIME2025 & 79.17  & 8.60  & 2.40  & 15.77\textsubscript{1.08x} & 26.82\textsubscript{1.83x} \\
      \bottomrule
      \end{tabular}%
    \label{tab:math_para_metric}%
  \end{minipage}
\end{table}%

As demonstrated in Table \ref{tab:math_scores} (reporting pass@1 scores based on Evalchemy \citep{Evalchemy},with AMC and AIME results representing means across 8 random seeds),
 ASPD exhibits performance gains of 12\%, 27.19\%, 16.67\%, 44.58\%, 37.5\% over
 \textbf{Ori} across MATH500, AMC23, GPQA, AIME24 and AIME25 respectively.
Notably, ASPD achieves acceleration ratios of 1.04-1.17 in TPS and 1.54-1.99 in P-TPS versus the \textbf{Seq} model, 
while maintaining performance within a range of -0.4\% to +5\%, demonstrating robust effectiveness.


\textbf{Parallelism Variation Across Tasks. }Compared to the results of other scenarios in Figure \ref{fig:token_speed_analysis}, 
mathematical tasks demonstrate relatively lower acceleration benefits (1.17x speedup in MATH500, 1.46x in RAG, and 1.82x in Vicuna).
In mathematical reasoning tasks, our parallel model contains reasoning processes with step-by-step deductions. 
These reasoning chains and strong inter-step dependencies lead to reduced DP,
which aligns with the parallel pattern shown in Figure \ref{fig:ourmethod} (30.6\% for mathematics versus approximately 68\% for other domains).
This reduction in parallelizable content consequently results in decreased Tokens-Per-Second (TPS) acceleration ratios.

\begin{figure}[htbp]
    \centering
    \begin{subfigure}[b]{0.48\textwidth}
        \centering
        \includegraphics[width=\textwidth]{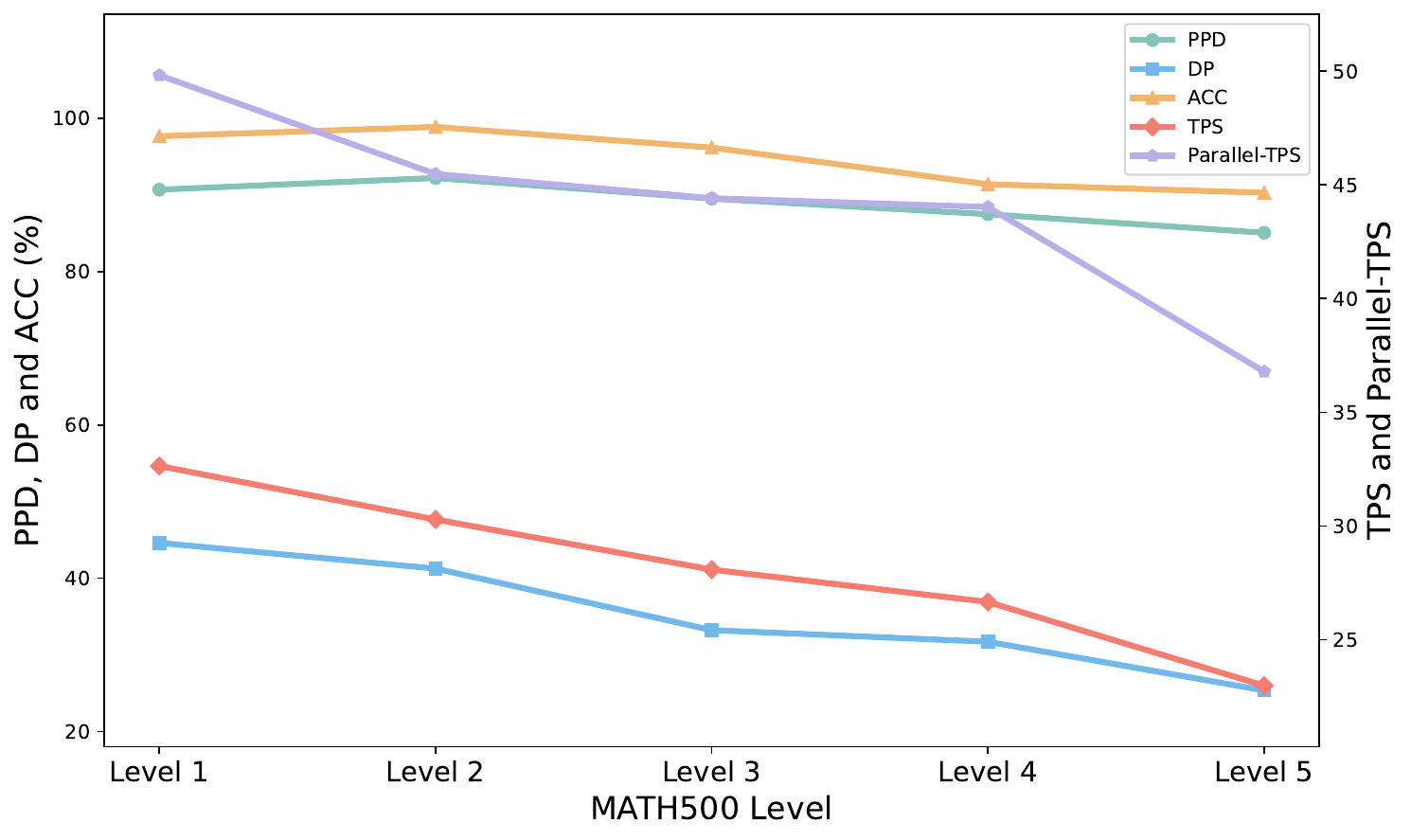}
        \caption{Parallelization Metrics vs. Problem Difficulty in MATH500.}
        \label{fig:math_500_details}
    \end{subfigure}
    \vspace{1em}
    \begin{subfigure}[b]{0.48\textwidth}
        \centering
        \includegraphics[width=\textwidth]{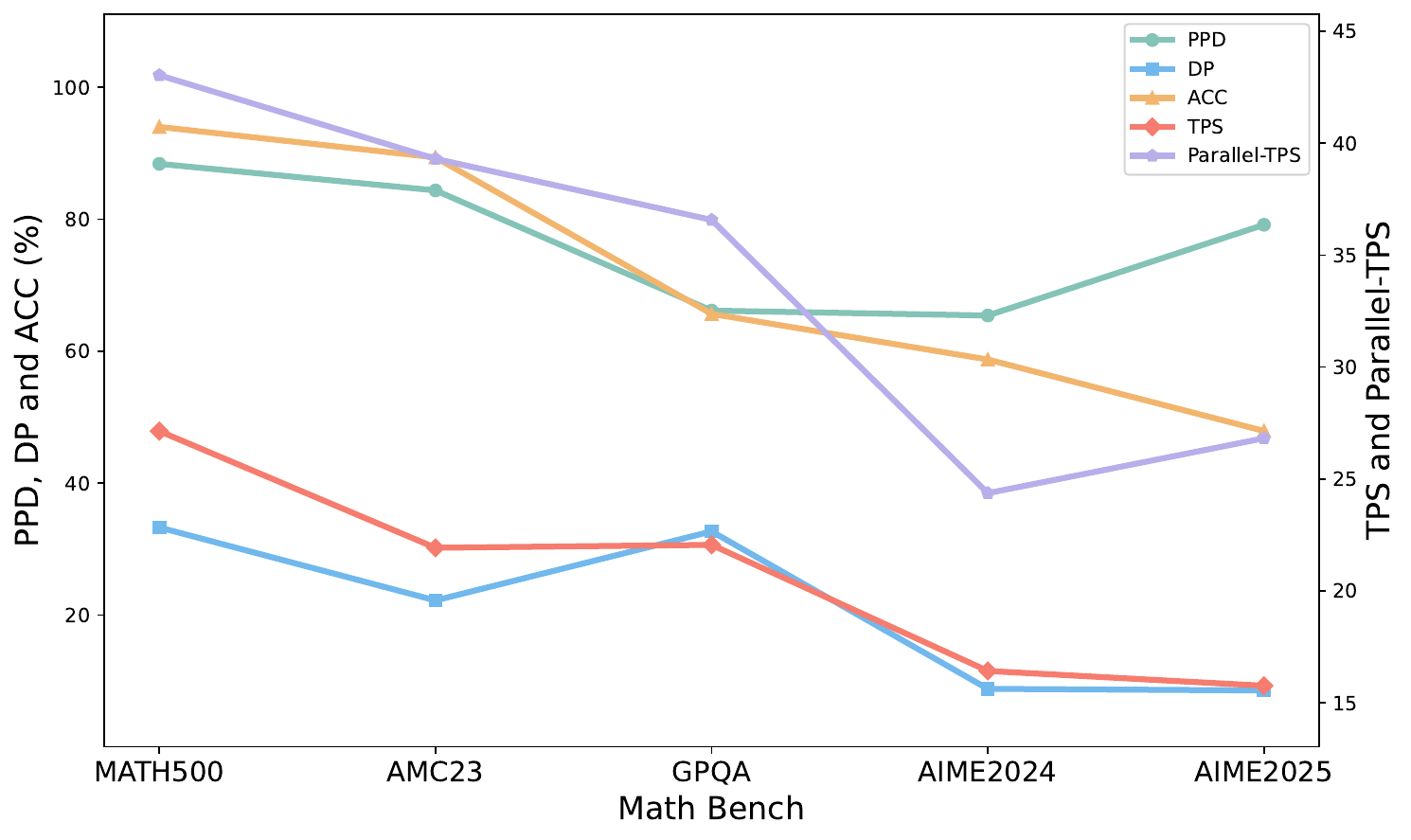}
        \caption{Parallelization Metrics across Math Benches.}
        \label{fig:mathbench_details}
    \end{subfigure}
    \caption{Analysis between Parallelization Metrics and Problem Difficulty.}
    \vspace{1em}
    \label{fig:data_analys_all}
\end{figure}

\textbf{Parallelism Variation Across Difficulty Levels. }
As shown in Table \ref{tab:math_para_metric}, the DP significantly decreases from 33.30\% in MATH500 to 8.6\% in AIME2025, with TPS decreasing from 1.17x to 1.08x.
This decline in parallelism indicates a strong correlation between performance degradation and difficulty level. 
Our detailed analysis of MATH500 in Figure \ref{fig:math_500_details} clearly shows that parallel efficiency metrics decrease as problem difficulty increases. 
This inverse relationship is consistently observed across various mathematical benchmarks in Figure \ref{fig:mathbench_details},
 where higher difficulty levels corresponds to lower DP and TPS values. 
Notably, the GPQA benchmark presents an interesting exception - its multiple-choice format naturally enables more parallel processing opportunities, 
leading to higher parallelization metrics despite comparable difficulty levels with other benchmarks.

\subsection{Ablation Study}

\subsubsection{Impact of Data Processing Pipeline. }
While autoregressive models inherently possess parallelizable characteristics in their responses, 
precisely identifying parallelizable components remains a significant challenge. 
As discussed in Section \ref{sec:para_data_pipe}, we must ensure semantic-level parallelization while maintaining branch independence and completeness. 
Table \ref{tab:datapipe} demonstrates the performance variations across different data processing methodologies. 
For our comparative analysis, we implemented APAR\citep{liu2024APAR} using their official codebase and PASTA\citep{jin2025learning} 
using the prompt provided in their paper (as their official code is not open-sourced). 
All construction processes utilized the Qwen3-235B-22A\citep{yang2025qwen3} model.

As shown in Table \ref{tab:datapipe},  APAR achieves only 1.11x speedup compared to the baseline, 
due to its rule-based parallelization approach which fails to trigger parallel responses in many scenarios. 
PASTA exhibits the lowest performance score of 4.98, 
primarily because its data processing pipeline lacks consideration for branch independence verification. 
In contrast, our method achieves optimal results in both effectiveness (score of 7.64) and efficiency (TPS of 104.21), 
demonstrating a 1.96x speedup while maintaining superior response quality.
\begin{table*}[htbp]
  \centering
  \begin{minipage}[t]{0.32\textwidth}
    \centering
    \caption{DataPipe Comparison}
    \begin{tabular}{c@{\hspace{2pt}}|@{\hspace{2pt}}c@{\hspace{2pt}}c}
    \toprule
    \bf Method & \bf Score & \bf TPS \\
    \midrule
    Baseline & \underline{6.21} & 53.19 \\
    APAR$^*$ & 5.81 & 59.25 \\
    PASTA$^\dagger$ & 4.98 & \textbf{106.83} \\
    \bf ASPD & \textbf{7.64} & \underline{104.21} \\
    \bottomrule
    \end{tabular}%
    \begin{tablenotes}
    \footnotesize
    
    \item{*} Implement with official codebase.\\
    \item{†} Implement with official prompt.
    \end{tablenotes}
    \label{tab:datapipe}%
  \end{minipage}%
  \hfill
  \begin{minipage}[t]{0.32\textwidth}
    \centering
    \caption{Attention Mask Study}
    \begin{tabular}{c@{\hspace{2pt}}c@{\hspace{2pt}}|@{\hspace{2pt}}c@{\hspace{2pt}}c}
    \toprule
    \bf PosId & \bf Attention & \bf Score & \bf TPS \\
    \midrule
    Seq & Shared & 4.64 & \textbf{110.30} \\
    \bf Seq & \bf Independent & \textbf{7.64} & \underline{104.21} \\
    Max & Shared & 3.70 & 86.96 \\
    Max & Independent & \underline{6.78} & 89.45 \\
    \bottomrule
    \end{tabular}%
    \label{tab:attention}%
  \end{minipage}%
  \hfill
  \begin{minipage}[t]{0.32\textwidth}
    \centering
    \caption{Position ID Study}
    \begin{tabular}{c@{\hspace{2pt}}|@{\hspace{2pt}}c@{\hspace{2pt}}c}
    \toprule
    \bf PosId  & \bf Score & \bf TPS \\
    \midrule
    Fix-32  & 6.09 & 59.11 \\
    Fix-128  & 4.99 & 72.44 \\
    Fix-512  & 4.24 & 61.00 \\
    Same-Max  & 6.78 & 89.45 \\
    Same-Re  & \underline{7.29} & \underline{95.24} \\
    \bf Same-Seq  & \textbf{7.64} & \textbf{104.21} \\
    \bottomrule
    \end{tabular}%
    \label{tab:position}%
  \end{minipage}%
\end{table*}%

\subsubsection{Impact of Mask Visibility. }
While  \citep{hsu2025group} and \citep{rodionov2025hogwild} have explored visible masks for agent collaboration, 
our approach differs fundamentally in its objective.
As demonstrated in Table \ref{tab:attention}, our empirical evaluation shows that branch-invisible masks consistently outperform visible masks across both sequential and max PositionID configurations. This architectural choice is motivated by our primary objective of decomposing responses into independent, self-contained parallel branches to achieve optimal computational efficiency.
The implementation of inter-branch visibility would compromise the critical independence property of parallel branches, potentially leading to content redundancy and semantic conflicts between branches, thereby degrading the overall response coherence and quality.
This empirical finding strongly validates our design decision to maintain strict branch isolation as an optimal strategy for parallel response generation.

\subsubsection{Position Encoding Paradigms. }
During parallel decoding, multiple branches decode simultaneously while remaining mutually invisible,
making the arrangement of position IDs particularly crucial. 
We abstract position encoding schemes into the following categories:

\textbf{Parallel Branches with Pre-allocated Fixed-Length Position IDs}. Pre-allocated fixed-length position ID intervals for each parallel branch, which are adopted by PASTA \citep{jin2025learning}. Our experiments show that with fixed intervals, model performance peaks at length 32 (score 6.09) but degrades significantly with larger intervals (4.24 at 512), suggesting an optimal interval matching the average parallel branch length in training data.

\textbf{Parallel Branches with Same Position IDs}. All parallel branches share identical position ID at each timestamp, with three variants:
\begin{itemize}
\item \textbf{Same-Max:} Uses maximum position ID across parallel branches for merging, achieving good performance (6.78) with moderate speed (89.45).

\item \textbf{Same-Rearrange(Same-Re):} Reorders position IDs sequentially post-decoding, requiring KV-cache re-prefill which introduces additional overhead.This implementation achieves suboptimal performance (7.29) and speed (95.24), representing a compromise between computational efficiency and generation quality.
\item \textbf{Same-Sequential(Same-Seq):} Assigns actual sequential position IDs, demonstrating optimal balance between performance (7.64) and speed (104.21).The performance degradation of Same-Re relative to Same-Seq can be attributed to the inconsistency between position IDs in the parallel phase and subsequent sequential phase during training. This discrepancy likely disrupts the model's optimization trajectory, ultimately resulting in suboptimal performance metrics.
\end{itemize}

As evidenced in Table \ref{tab:position}, 
our adopted Same-Sequential strategy emerges as the most effective strategy, 
outperforming both Fixed and Same-Max variants in terms of quality and efficiency. 
This suggests that maintaining natural position ordering while sharing timestamps across parallel branches optimally preserves the model's positional understanding.

\section{Conclusion}
In this work, we present \textbf{ASPD}: The \textbf{A}daptive \textbf{S}erial-\textbf{P}aralle \textbf{D}ecoding framework for efficient hybird decoding within auto regressive large language models.
The proposed method introduces a non-invasive parallel data transformation pipeline and internal parallelization with branch-invisible attention masks and shared position IDs,
achieving substantial latency reduction while maintaining response quality compared to traditional autoregressive LLMs.
 
 Our extensive experiments demonstrate state-of-the-art performance across various benchmarks including general tasks, 
  retrieval-augmented generation, and mathematical reasoning. 
Furthermore, we establish a novel paradigm for parallel decoding that eliminates external overheads from batching, threading or re-prefill between serial-parallel switching,
   providing valuable insights for future research in efficient LLM inference. 
   The strong empirical results and theoretical contributions of ASPD suggest promising applications in latency-critical scenarios.

\section{Limitations and Future Work}
Our approach focuses on semantic-level parallelism through concurrent generation of independent response segments,
while speculative decoding achieves token-level parallelism via predictive token generation with sequential verification.
These orthogonal yet complementary approaches present opportunities for future research to combine both paradigms for enhanced acceleration.

Moreover, our parallelization modifications exhibit significant potential for seamless integration with mainstream inference frameworks, including
vLLM\citep{kwon2023efficient} and SGLang\citep{zheng2024sglang}, promising substantial enhancements in acceleration performance.

Within our parallel data production pipeline, we have established several key evaluation metrics, including TPS, DP, and ABN.
These quantitative indicators demonstrate significant potential for integration into reinforcement learning frameworks,
where they could effectively guide model optimization towards achieving enhanced parallelism while maintaining or improving overall performance.
\section*{Contributions}

\paragraph{Authors}
Keyu Chen$^{*}$ \textsuperscript{\rm 1}\quad Zhifeng Shen$^{*}$ \textsuperscript{\rm 1}\quad Daohai Yu$^{*}$ \textsuperscript{\rm 1,2}
\quad Haoqian Wu \textsuperscript{\rm 1}\quad Wei Wen \textsuperscript{\rm 1} \quad Jianfeng He \textsuperscript{\rm 1}
\quad Ruizhi Qiao \textsuperscript{\rm 1} \quad Xing Sun \textsuperscript{\rm 1}\quad 



\paragraph{Affiliations}
\begin{minipage}[t]{\linewidth}
\textsuperscript{1}Tencent YouTu Lab\\[2pt]
\textsuperscript{2}Key Laboratory of Multimedia Trusted Perception and Efficient Computing,\\
\hspace*{0.5em}Ministry of Education of China,Xiamen University, Xiamen 361005, P.R. China
\end{minipage}

\blfootnote{$^{*}$ Equal Contribution}


\setcitestyle{numbers,square}
\setcitestyle{square,numbers,comma}
\bibliography{youtu_bib}

\end{document}